\documentclass[conference]{IEEEtran}
\IEEEoverridecommandlockouts
\usepackage{multirow}
\usepackage{bbding}

\usepackage{makecell}
\usepackage{circledsteps}
\usepackage{tikz}
\newcommand*\circled[1]{\tikz[baseline=(char.base)]{
            \node[shape=circle,draw,inner sep=2pt] (char) {#1};}}

\usepackage{cite}
\usepackage{amsmath,amssymb,amsfonts}
\usepackage{algorithmic}
\usepackage{graphicx}
\usepackage{textcomp}
\usepackage{xcolor}
 \usepackage{todonotes}
\def\BibTeX{{\rm B\kern-.05em{\sc i\kern-.025em b}\kern-.08em
    T\kern-.1667em\lower.7ex\hbox{E}\kern-.125emX}}
\begin{document}

\title{Power-Performance Characterization of \\ TinyML Systems
\thanks{\textit{Proceedings of the ICCD Conference}, Lake Tahoe, USA, 2022. Copyright 2022 by the author(s).}
}

\author{\IEEEauthorblockN{Yujie Zhang, Dhananjaya Wijerathne, Zhaoying Li, Tulika Mitra}

\IEEEauthorblockA{\textit{School of Computing, National University of Singapore} \\
\{zyujie, dmd, zhaoying, tulika\}@comp.nus.edu.sg}
}

\maketitle

\begin{abstract}
    TinyML systems are enabling machine learning (ML) inference at the edge. However, there exists little quantitative analysis of such systems. This paper presents a systematic performance and power characterization of diverse TinyML applications on microcontrollers (MCUs), spanning neural network models, software libraries, operating systems, and hardware architectures. We focus on the impact of the multiple layers of abstraction that provide higher programmability at the expense of performance and energy efficiency. We propose a model to estimate the costs of different abstraction layers and make recommendations for minimizing those costs. Our findings can help designers with Neural Architecture Search (NAS) and CNN inference optimization on edge devices.
\end{abstract}



\section{Introduction}
Machine learning (ML) inference in the cloud is the most prevalent approach today. However, inference on edge devices closer to the data source is becoming increasingly important due to security and privacy concerns. Moreover, processing the data at the edge saves network bandwidth and energy. Finally, some applications require a real-time response that cannot be guaranteed with unpredictable network delay and availability. TinyML~\cite{tinyml.org} is a special class of inference on ultra-resource-constrained edge devices that can achieve data collection and processing in real-time with very little power~\cite{banbury2020benchmarking}.

There exists a large body of work on power and performance analysis of ML training systems~\cite{wang2019benchmarking,wang2020benchmarking} to provide insights into future system design. There is, however, very little quantitative analysis of the capabilities of existing TinyML systems. Even though ~\cite{banbury2021micronets,heim2021measuring,yazdanbakhsh2021evaluation} analyze deep neural network (DNN) implementation on MCU or Edge TPU, they do not adopt a full-stack view to study the limitations of DNN implementation on edge devices. A systematic and fine-grained power-performance characterization can help us identify the bottleneck of the current TinyML system and improve the performance of the edge device.

More importantly, various TinyML optimizations complicate the performance analysis. Severely resource-constrained devices require significant neural network models, software, and hardware-level optimizations to support complex ML applications. For instance, model-level optimizations, such as weight pruning and quantization, reduce memory footprint and inference computation complexity with negligible loss of accuracy. 
Architecture-specific libraries and compilers like TensorFlow Lite Micro (TFLM)~\cite{david2021tensorflow} framework for microcontrollers, and the CMSIS-NN library~\cite{lai2018cmsis} for ARM Cortex-M processors influence inference latency. Moreover, microcontroller architecture and platform details also substantially impact the power-performance behavior of ML inference.


While each level (model, framework, compiler/library, architecture) can significantly affect the performance of TinyML applications, their subtle interactions also have considerable importance due to the ultra-resource-constrained nature of these devices.
Thus, the final inference performance is simultaneously determined by the whole stack, from model structure, compilation and software optimization, and architecture. 
For instance, matrix multiplication implementation of DNN layers using single instruction multiple data (SIMD) instructions can significantly accelerate model inference and improve energy efficiency. In this case, latency/energy benefit is obtained from hardware-specific model design or optimization.

We provide a systematic power and performance analysis of TinyML systems, considering deep neural networks, the software library, the operating system, and hardware architectures. This analysis identifies the bottlenecks of the existing systems and indicates how to optimize the system further.

We evaluate five representative TinyML applications \cite{warden2019tinyml,banbury2021mlperf} on different MCU platforms regarding inference latency, power efficiency, and energy consumption. We explore the impact of abstraction layers (operating system and ML framework) on model inference. Next, we investigate the general matrix multiplication (GEMM) implementation in software libraries and suggest strategies to optimize the corner cases. We study how hardware characteristics affect inference performance and provide insights on supporting larger NN models in ML frameworks. Finally, we propose a model to predict the ML inference performance on MCUs considering the impact of the full stack. Our study provides insights into future system optimizations and can be easily integrated into the platform-specific neural architecture search (NAS) process, for example. 


\section{Related Work}
\label{section:related_work}
Table \ref{tab:related_work}  shows a comparison of related works in system benchmarking for ML applications with our study. 

There is much systematic analysis of DNN training platforms to identify the weaknesses and strengths of various targeted hardware architectures under different model structures. In terms of ML inference on edge devices, there is very little quantitative analysis, which to some extent impedes TinyML system development. Banbury et al.~\cite{banbury2020benchmarking} point out that the main difficulty is the lack of a universally acknowledged benchmark, and they present four TinyMLPerf benchmarks. They provide a way to benchmark TinyML systems but do not analyze the system behaviors. 

Based on the observation on MCU platforms that model latency is proportional to model operation count, Banbury et al.~\cite{banbury2021micronets} propose MicroNet models and achieve state-of-the-art performance for three TinyMLPerf benchmark tasks. Heim et al.~\cite{heim2021measuring} show that perceptible performance metrics, like inference latency and energy consumption, should be a viable proxy for model design and present a toolchain for TensorFlow Lite (TFLite)~\cite{TensorFlowLite} model implementation on MCU platforms. Yazdanbakhsh et al.~\cite{yazdanbakhsh2021evaluation} evaluate three Edge TPUs and present a learned ML model to predict inference latency for  convolutional kernels. While these works only focus on MCUs or Edge TPU, Baller et al.~\cite{baller2021deepedgebench} provide an inclusive analysis of deep learning inference performance on various edge devices, like GPU, NPU, and MCU. Their analysis mainly focuses on model latency/energy with or without utilizing AI units. Wang et al.~\cite{wang2020neural} provide quantitative power-performance analysis of different components on mobile SoCs for deep learning inference and propose synergistic engagement of all the components concurrently to improve throughput~\cite{wang2019high}.  Compared with these works, we systematically analyze the overhead and limitation of model inference on MCUs from a full-stack perspective,  discuss how to improve the performance and provide a prediction model.


\vspace*{-0.1in}
\begin{table}[htbp]
    \centering
    \caption[Summary of state-of-the-art]{Summary of state-of-the-art}
    \resizebox{\linewidth}{!}{
        \begin{tabular}{ccc} 
        \hline
            &\textbf{Platforms}&\textbf{Highlight}\\
            \hline
            \cite{banbury2021micronets}&MCU&\makecell[l]{State-of-the-art MicroNets based on differentiable\\NAS for three TinyML tasks.}\\
            \hline
            \cite{heim2021measuring}&MCU&\makecell[l]{A toolchain of TFLite model implementation\\on MCUs using Mbed OS for NAS.}\\
            \hline
            \cite{yazdanbakhsh2021evaluation}&Edge TPU&\makecell[l]{1. Microarchitectural insights of Edge TPUs.\\2. Prediction model for faster evaluation of\\accelerators.}\\
            \hline
            \cite{baller2021deepedgebench}&\makecell[l]{Four SoC,\\one MCU}&\makecell[l]{Performance comparison of four SoCs and one\\MCU for different models and frameworks.}\\
            \hline
            Ours&MCU&\makecell[l]{Recommendation to overcome the performance\\costs from a full-stack view.}\\
        \hline
        \end{tabular}
    }
    \label{tab:related_work}        
\end{table}
\vspace*{-0.1in}

\section{Experimental Setup}
\label{section:ExperimentSetUP}
To evaluate the model inference performance on edge devices, we select five representative TinyML applications and evaluate them on multiple MCU platforms. 

\begin{figure}[htbp]
    \centerline{\includegraphics[width=\linewidth, height=0.35\linewidth]{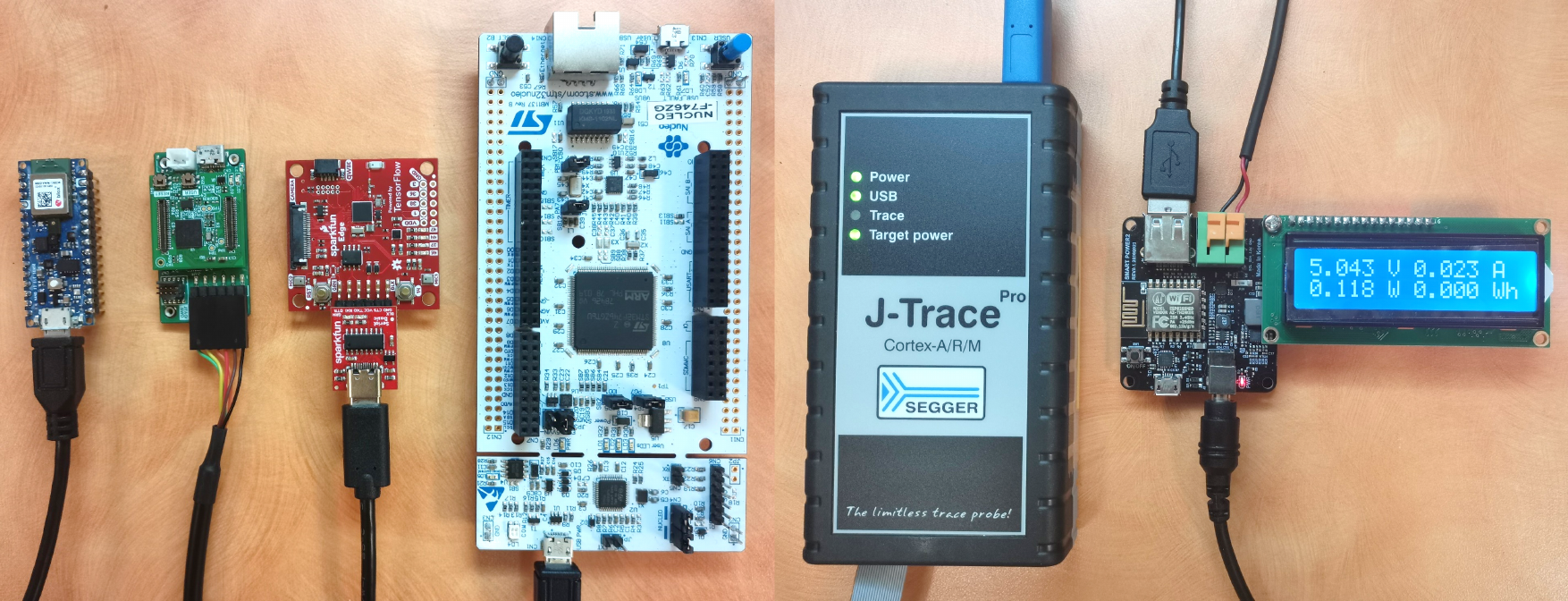}}
    \caption{Picture of Arduino Nano 33 BLE Sense, Nucleo-64 STM32L476RG, SparkFun Edge Apollo3 Blue, Nucleo-144 STM32F746ZG, J-Trace Pro Debug Probe, and SmartPower2 (from left to right).}
    \label{fig:MCUs}
\end{figure}

\textbf{MCU Platforms.} Considering the compute capability and memory availability, we select four MCU platforms in Table \ref{tab:MCU_architecture_information} for the evaluation. The mature toolchains supporting model inference on these MCUs also facilitate our evaluation.
We adopt Arduino Nano 33 BLE Sense~\cite{ArduinoNano}, SparkFun Edge Apollo3 Blue~\cite{SparkFunEdge}, Nucleo-144 STM32F746ZG~\cite{STM32F746ZG}, and Nucleo-64 STM32L476RG~\cite{STM32L476RG} (Fig.~\ref{fig:MCUs}). Their 32-bit ARM processor cores, Cortex-M4F or Cortex-M7F, are designed for cost and energy-efficient micro-controllers. Cortex-M7F can dual issue specific pairs of ALU instructions to achieve better performance. The devices vary in clock speeds, SRAM sizes, etc., which contributes to a compelling study of how hardware architectures affect inference performance. We will use their acronyms in Table \ref{tab:MCU_architecture_information} (MCU-S, MCU-M, MCU-B, MCU-D) to represent them for simplicity in the rest of the paper. 

\vspace*{-0.1in}
\begin{table}[htbp]
    \centering
    \caption{MCU Architecture information}
    \resizebox{\linewidth}{!}{
        \begin{tabular}{lcccc} 
        \hline
            &\makecell{Arduino Nano\\33 BLE Sense}&\makecell{Nucleo-64\\STM32L476RG}&\makecell{SparkFun\\Edge}&\makecell{Nucleo-144\\STM32F746ZG}\\
            \hline
            Acronym&\textbf{MCU-S}&\textbf{MCU-M}&\textbf{MCU-B}&\textbf{MCU-D}\\
            \hline
            Processor&Cortex-M4F&Cortex-M4F&Cortex-M4F&Cortex-M7F\\
            \hline
            Frequency&64 MHz&80 MHz&96 MHz&216 MHz\\
            \hline
            Bandwidth&67.3 MB/s&86.9 MB/s&88.6 MB/s&251.6 MB/s\\
            \hline
            \centering{SRAM}&256 KB&128 KB&384 KB&320 KB\\
            \hline
            \centering{Flash}&1 MB&1 MB&1 MB&1 MB\\
        \hline
        \end{tabular}
    }
    \label{tab:MCU_architecture_information}        
\end{table}

\textbf{TinyML Benchmarks.} We use five benchmarks~\cite{warden2019tinyml, banbury2021mlperf} presented in Table \ref{tab:model_description}. These applications vary in input types, model structures, and computation intensity. Evaluating them on edge devices facilitates understanding the fine-grained impact of model structures on inference characteristics.

\begin{table*}[htbp]
    \centering
    \caption{Model description of main benchmarking applications}
        \begin{tabular}{p{1.7cm}p{2.7cm}p{2.4cm}p{2.7cm}p{3cm}p{3cm}} 
        \hline 
            &\textbf{Micro Speech}&\textbf{Person Detection}&\textbf{Keyword Spotting}&\textbf{Image Classification}&\textbf{Anomaly Detection}\\
            \hline
            \textbf{Description}&Recognizing two wake words&Recognizing Person&10 keyword spotting —— "No", "Up",...&Small image classification&Detecting anomalies in machine operating sounds\\
            \hline
             \textbf{Model}&——&MobileNet~\cite{sandler2018mobilenetv2}&DS-CNN~\cite{zhang2017hello}&ResNet~\cite{he2016deep}&Deep AutoEncoder\\
            \hline
            \textbf{Layers}&Conv+FC&Conv+13 DSC+FC&Conv+4 DSC+FC&9 Conv+FC&10 FC\\
            \hline
            \textbf{Dataset}&Speech Commands&COCO&Speech Commands&Cifar10&ToyADMOS\\            
            \hline
            \textbf{\#MACC}&336k&7158k&2657k&12502k&264k\\
            \hline
            \textbf{OPs}&676k&15029k&5541k&25271k&538k\\            
            \hline
            \textbf{TFLite Model}& 18.7 KB & 300.6 KB & 59.6 KB & 98.5 KB & 277.0 KB\\       
        \hline
        \end{tabular}
        \label{tab:model_description}        
\end{table*}

\textbf{Model Implementation.} We use TFLite to obtain the quantized int8 version. Next, TensorFlow Lite Micro (TFLM)~\cite{david2021tensorflow} and CMSIS-NN library are adopted to run these applications on MCUs. TFLM does not need operating system support to run complex TinyML applications on resource-constrained MCUs, reducing the memory requirement. CMSIS-NN kernels leverage SIMD instructions to implement matrix multiplication, improving the inference latency and energy.  

\textbf{Experimental Measurement.} 
We use Mbed Timer API and clock registers to measure inference latency on Mbed OS and Bare Metal, respectively. We use SmartPower2~\cite{SmartPower2} to record the average power during model execution and then multiply it with inference time to obtain the energy. We use J-Trace Pro Debug Probe in ARM Keil MDK Professional Edition to trace MCU processor registers (e.g., CYCCNT register) and count the instructions executed during model evaluation. 


\section{Impact of Abstraction Layers}
\label{section:abstractionLayer}
We first explore the impact of multiple abstraction layers of the operating system and DNN framework on model inference performance on MCUs. The real-time operating system (RTOS) on embedded devices provides an interface between hardware peripherals and software programs. With multiple abstractions and implementation of Application Programming Interfaces (APIs), it greatly facilitates neural network inference programming. However, redundant support of APIs, like driver APIs and platform APIs, unavoidably causes application performance  and energy efficiency degradation. Similarly, the deep learning framework provides a unified way for neural network definition, interpretation, and execution, to improve program portability and generalization on different devices. However, programmability and generalization lead to higher inference latency and energy inefficiency. 

We use {\em Micro Speech} as an example to show the impact of the above abstractions. {\em Micro Speech} consists of typical convolution and fully connected layers. 
Table \ref{tab:overhead_analysis} summarizes the characteristics of different implementation choices on MCU-M: \circled{1} naive model implementation in C without OS, DNN optimization libraries, and framework support, \circled{2}  C code with CMSIS-NN optimization, \circled{3} C code with CMSIS-NN optimization on Mbed OS, \circled{4} C code with CMSIS-NN optimization on simplified Mbed OS, and \circled{5} TFLM framework with CMSIS-NN optimization on Mbed OS. 

For implementation choice \circled{1}, we implement the model inference in pure C code and execute it bare-metal, i.e., we do not use any deep learning library, framework, or rely on an operating system. For implementation choice \circled{2}, we utilize CMSIS-NN kernels, which provide optimized implementation of convolution and fully connected operations. The inclusion of optimized kernels leads to around 10x improvement in runtime and energy efficiency. The reason is that CMSIS-NN utilizes SIMD instructions and matrix multiplication to reduce memory access and enable data reuse, significantly improving the model inference performance. 

\newcommand{\tabincell}[2]{\begin{tabular}{@{}#1@{}}#2\end{tabular}}  

\begin{table}[htbp]
    \centering
    \caption{Overhead analysis of abstraction layers based on\\  MicroSpeech on MCU-M}
    \resizebox{\linewidth}{!}{
        \begin{tabular}{lrrrr} 
        \hline
            &\tabincell{c}{Latency\\(ms)}&\tabincell{c}{SRAM\\(KB)}&\tabincell{c}{Binary\\(KB)}&\tabincell{c}{Energy\\(mJ)}\\
            \hline
            {\footnotesize1) }Unoptimized code+Bare-Metal&488.3&6.6&37&54.2\\        
            \hline
            {\footnotesize2) }CMSIS-NN+Bare-Metal&46.2&4.1&28&5.1\\
            \hline
            {\footnotesize3) }CMSIS-NN+Mbed OS&47.5&8.1&83&8.4\\
            \hline
            {\footnotesize4) }CMSIS-NN+Simpl. Mbed OS&47.5&7.7&53&5.6\\              
            \hline
            {\footnotesize5) }TFLM+CMSIS-NN+Mbed OS&48.9&11.8&99&9.5\\
        \hline 
        \end{tabular}
    }
    \label{tab:overhead_analysis}        
\end{table}

Next, we explore the impact of the operating system in implementation choice \circled{3} where we run the application with CMSIS-NN optimization on Mbed OS. Although Mbed OS support does not affect the latency much, adding OS support increases SRAM and Flash memory requirements by approximately 98\% and 196\%, respectively. Besides, as bare metal implementation does not need extra peripheral settings of Mbed OS, the latter also requires  65\% additional energy. Thus, without extra peripherals and abstractions layer support, bare metal implementation allows small MCU platforms to support large models with less energy. To quantify the overhead of unnecessary peripheral support, we remove all unnecessary peripheral API support of Mbed OS to obtain a simplified operating system in implementation choice \circled{4}. Compared with full-fledged Mbed OS, model inference on simplified Mbed OS achieves considerable improvement in memory and energy and offers a good compromise between programmability and efficiency. 

Finally, to evaluate the overhead of deep learning framework on model inference, we use TFLM for model representation and interpretation instead of direct implementation in C for implementation choice \circled{5}. TFLM on Mbed OS  causes around 3\% increase in inference time, energy, and memory requirements. Clearly, the advantages of TFLM abstraction far outweigh the overhead on performance efficiency.

\textbf{Conclusion and Recommendation.} Our study on the impact of operating system and ML framework on inference latency/energy and memory requirements quantifies the trade-off between programmability and efficiency. Specifically,  the OS causes considerable overhead in terms of memory and energy. However, it can be mitigated to a large extent by carefully removing the support for redundant peripherals from the OS even if it requires some engineering effort. On the other hand, the deep learning framework itself adds little overhead and substantially reduces the designer's effort. 

\section{Optimization Library Limitation}
\label{section:optimization_library}
We now study a widely used and representative deep learning library CMSIS-NN on MCUs to analyze its performance limitations due to the interaction between the optimizations and the underlying architecture. Based on the study, we provide suggestions for NAS to avoid resource under-utilization due to inappropriate model parameter settings and further optimizations to utilize the limited resource. 

\subsection{SIMD Optimization Usage}
\label{subsection:SIMD_optimization}
Matrix multiplication is the most important and computationally intensive kernel for neural network inference~\cite{warden2015gemm}. Currently, many industry-standard backend NN libraries, such as CMSIS-NN, CMix-NN~\cite{capotondi2020cmix}, X-CUBE-AI~\cite{X-CUBE-AI}, and PULP-NN~\cite{garofalo2020pulp}, utilize matrix multiplication to enable ML on resource-limited edge devices. In the CMSIS-NN implementation, matrix multiplication kernels are used for most neural network layers (convolution, depthwise separable convolution, and fully connected layers). However, this conversion is not always suitable. 



\begin{figure}[htbp]
    \centerline{\includegraphics[width=0.8\linewidth,height=0.4\linewidth]{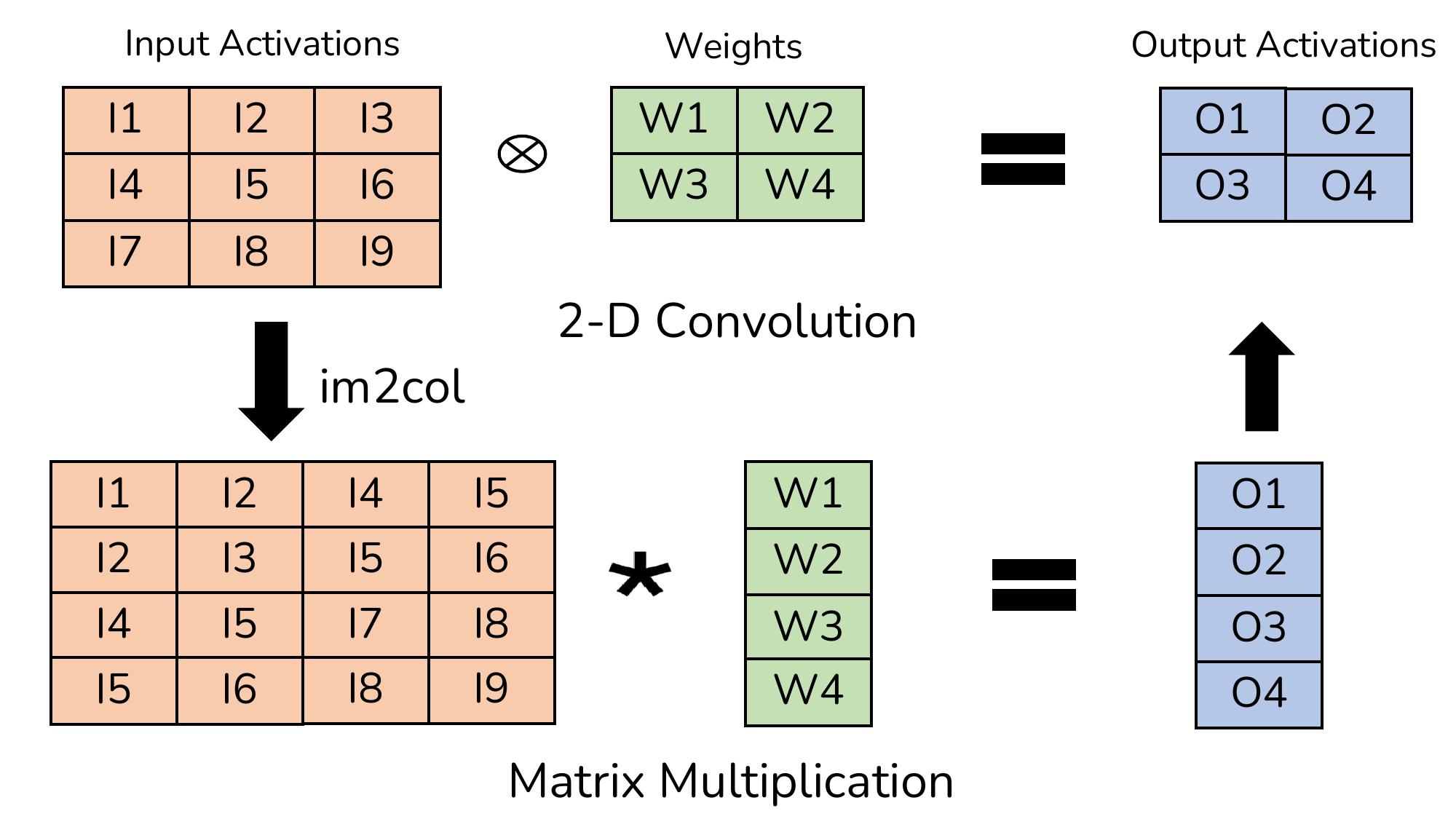}}
    \caption{An example of 2-D convolution implementation.}
    \label{fig:conv_to_gemm}
\end{figure}

Fig.~\ref{fig:conv_to_gemm} shows how a 2D convolution operation with 2x2 filter weights can be lowered to matrix-vector multiplication.
Input activations are replicated to create a matrix where each row corresponds to the elements in one convolution window (2x2 window).
Therefore, one output activation can be generated by performing the dot product of the flattened filter weights with every row of the matrix.
Similarly, 3D convolution and depthwise separable convolution (DSC) can be converted to general matrix multiplication (GEMM). 
Input activation replication is also known as input expanding, and this transformation is performed using the image-to-column (IM2COL) process. 

The actual implementation in CMSIS-NN is more complicated. For example, in the implementation of convolution operator, CMSIS-NN uses partial IM2COL, in which only a limited number of input columns are expanded to avoid high memory overhead. The matrix-multiplication kernel used by convolutions is implemented with a 2x2 kernel, which is carefully designed to fit the limited architectural registers of Cortex-M processors. Its computation is based on the dedicated Multiply-and-Accumulate (MAC) instruction \textit{\_\_SMLAD}. Computations that cannot utilize matrix-multiplication kernels due to tricky data alignment from inappropriate parameter settings are implemented through regular ALU instructions \textit{MLA} and \textit{ADD}. Compared with SIMD instructions, regular ALU instructions need more time for MAC computation. More extensive usage of SIMD instructions in convolution corresponds to improved runtime. 

 
The following example layers of {\em Person Detection} application show this limitation of CMSIS-NN GEMM-based optimization. We list \#MACC, \#MEM, and latency of four Conv1x1 layers (A-D) in Table \ref{tab:conv1x1_time_MCU}. \#MACC is the number of MACC operations while \#MEM denotes the number of memory access operations. Both numbers are calculated based on the model description. As the executable memory needed by this application exceeds the SRAM size of MCU-M, we only show experimental results on other three MCUs.  

As shown in Table \ref{tab:conv1x1_time_MCU}, layer B needs to execute nearly twice memory access and MACC operations compared to layer A. However, B's latency is only 0.69x times that of A's on MCU-B. On the other two platforms, this ratio is even lower. The small number of input channels of layer A fails to take advantage of GEMM conversion. Most of its computation has to be handled by normal ALU instructions without any SIMD optimization. In this case, layer A needs more inference time than some layers like layer B with large input channels, even though they have higher \#MACC and \#MEM. A similar observation applies to layer C and layer D. 

\begin{table}[htbp]
    \centering
    \caption{Conv1x1 layer latency on MCU platforms}
        \begin{tabular}{crrrrr} 
        \hline
            &\textbf{\#MACC}&\textbf{\#MEM}&\textbf{MCU-S}&\textbf{MCU-B}&\textbf{MCU-D}\\
            \hline
            Layer A&295k&627k&66 ms&31 ms&11 ms\\
            \hline
            Layer B&590k&1198k&40 ms&21 ms&6 ms\\
            \hline
            Ratio(B/A)&2x&1.91x&\textbf{0.61x}&\textbf{0.69x}&\textbf{0.53x}\\
            \hline
            Layer C&295k&608k&28 ms&15 ms&5 ms\\
            \hline
            Layer D&590k&1184k&31 ms&16 ms&4 ms\\
            \hline
            Ratio(D/C)&2x&1.95x&\textbf{1.09x}&\textbf{1.04x}&\textbf{0.92x}\\
        \hline
        \end{tabular}
        \label{tab:conv1x1_time_MCU}        
\end{table}

\textbf{Conclusion and Recommendation.} Due to constant matrix multiplication kernel size, some computation of layers with inappropriate shapes has to be completed using normal instructions. However, naive computation without matrix multiplication kernels support increases inference latency due to slower MAC calculation and memory access. Careful parameter setting to enable converting all model computation to matrix multiplication can greatly facilitate inference performance.


\subsection{Kernel Implementation Overheads}
\label{subsection:direct_convolution}

In this section, we investigate the overheads of CMSIS-NN kernel implementations for convolution, fully connected, and depthwise separable convolution layers.

\begin{figure}[htbp]
    \centerline{\includegraphics[width=\linewidth]{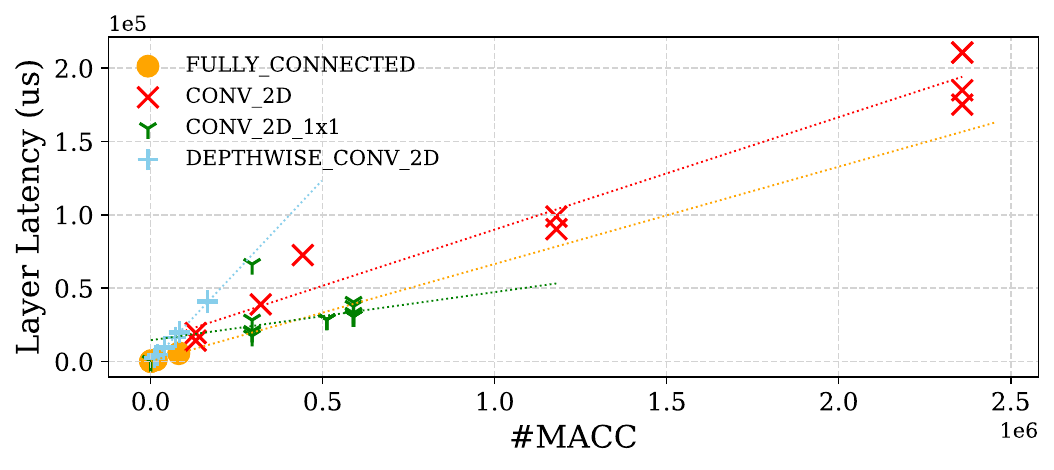}}
    \caption{Layer Latency on MCU-S.}
    \label{fig:layer_latency_MCU-S}
\end{figure}

Fig.~\ref{fig:layer_latency_MCU-S} shows the inference latency of layers from five TinyML applications in Table \ref{tab:model_description} on MCU-S.
The X-axis represents the number of MACC operations in each layer.  
Convolution layers typically require fewer parameters (weights and biases) than fully connected layers.
Therefore operational intensity~\cite{williams2009roofline} (number of operations per memory access) of a fully connected layer is smaller than that of a convolutional layer. This indicates that if both the fully connected and convolution layers have the same number of MACC, the fully connected layer will have more memory accesses and then take more inference time.
However, as shown in Fig. \ref{fig:layer_latency_MCU-S}, the latency of the convolutional layer becomes higher than that of the fully connected layer with the same number of MACC operations. 
This inconsistency is also caused by the GEMM conversion in the CMSIS-NN kernel implementations. 

As mentioned earlier, most NN libraries including CMSIS-NN, lower all three layers (convolution, fully connected, and depthwise separable convolution) to matrix multiplication kernels.
Convolution and DSC layers require the IM2COL process for input expansion.
However, as fully connected layer implementation is inherently a matrix multiplication, it can be converted to matrix multiplication without input expansion. 
GEMM conversion in CMSIS-NN kernels generates overhead for convolution layer inference latency and memory footprint. 
The operational intensity of convolution decreases due to input expansion, which makes it easily limited by memory bandwidth. Moreover, reordering and expanding partial inputs also incur more time and memory costs than direct convolution. 
Even with partial IM2COL, extra memory cost is unavoidable. 


With the same number of MACC operations, depthwise convolutions cause more overhead in the IM2COL process than convolutions. Although GEMM conversion decreases the operational intensity of the depthwise convolutions and the convolutions to a different extent, depthwise convolutions still exhibit higher latency per MACC than convolutions.

\begin{figure}[htbp]
    \centerline{\includegraphics[width=\linewidth]{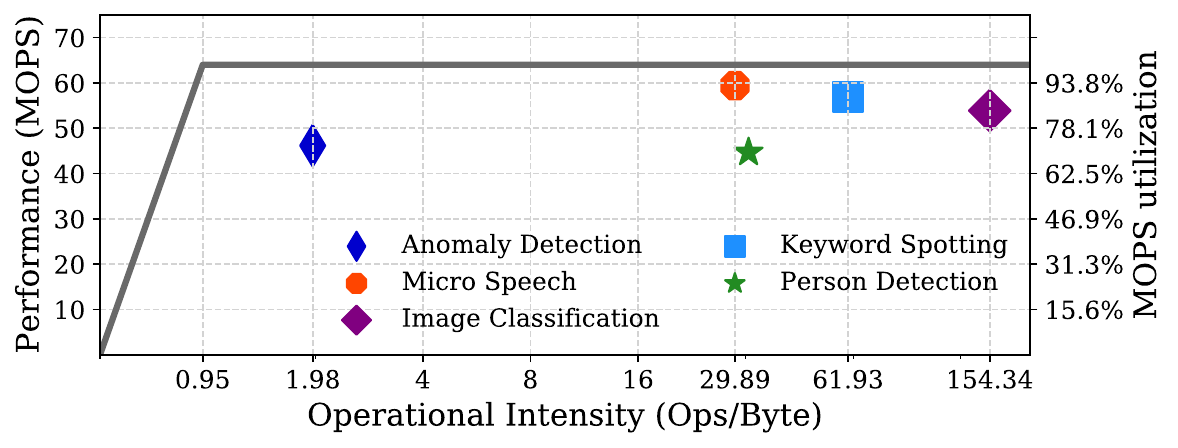}}
    \caption{Roofline model of MCU-S.}
    \label{fig:roofline_MCU-S}
\end{figure}

Next, we calculate the operational intensity of five applications based on model description and draw the roofline model of MCU-S in Fig.~\ref{fig:roofline_MCU-S}. The Y-axis on the left represents application performance (million operations per second, MOPS) and the right one denotes MOPS utilization, the ratio of application MOPS to peak MOPS. We utilize application's MOPS utilization to measure the computation capacity utilization efficiency on the platform. 
Based on the roofline model, these applications are all computation-bound. However, due to IM2COL overhead and decreased operational intensity from GEMM conversion, most  applications, especially {\em Person Detection}, cannot fully utilize platform compute capability. For instance, the MOPS utilization of {\em Person Detection} and {\em Image Classification} are only 69.8\% and 84.2\%, respectively. Besides,  {\em Anomaly Detection} consisting of fully connected layers achieves only 72.1\% MOPS utilization, because of the incomplete utilization of SIMD optimization as mentioned in Section \ref{subsection:SIMD_optimization}. The layer inference computation without matrix-multiplication kernels requires more load instructions with one-byte inputs or weights access, significantly wasting memory bandwidth. 

In order to investigate how GEMM conversion affects energy efficiency, we compare latency/energy for five applications on MCU-S, as shown in Fig.~\ref{fig:power-performance-numbers}.  The applications with  more  latency  consume more energy due to the near-constant power during inference. Therefore, the overhead caused by GEMM conversion at runtime impedes energy efficiency.

\textbf{Recommendation.} In view of the cost of GEMM conversion in NN library  kernels,  direct convolution implementation with SIMD optimization should be explored as a promising alternative. For example, Zhang et al.~\cite{zhang2018high} demonstrate that an implementation of direct convolution achieves better performance than existing state-of-the-art CPU-based convolution implementations. Gural et al.~\cite{gural2019memory} propose memory-optimal direct convolutions by performing computations in place on resource-constrained devices. 


\section{Architectural Impact}
\label{section:architectural_impact}
We now characterize the influence of hardware characteristics on power and performance in ML inference on MCUs. 

\subsection{Energy Efficiency}
\begin{figure}[htbp]
    \centerline{\includegraphics[width=\linewidth]{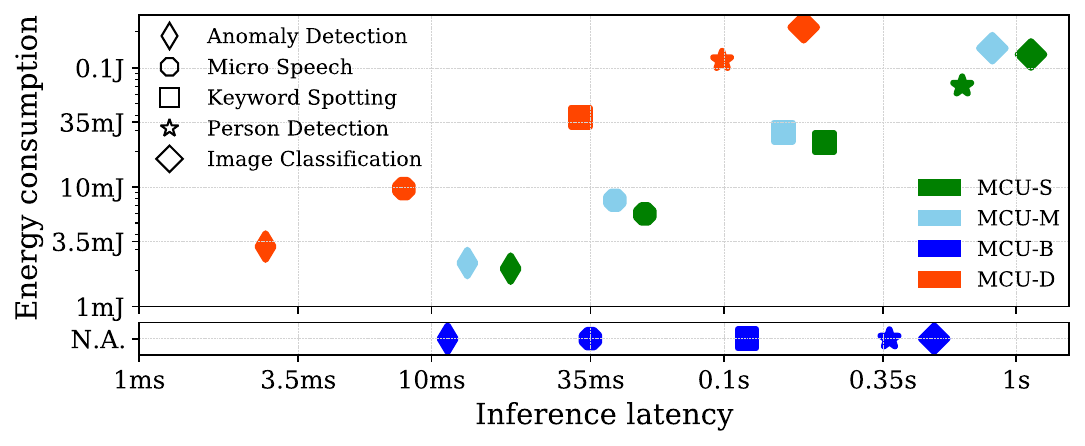}}
    \caption{Application latency/energy numbers on MCUs.}
    \label{fig:power-performance-numbers}
\end{figure}

\begin{figure}[htbp]
    \centerline{\includegraphics[width=\linewidth]{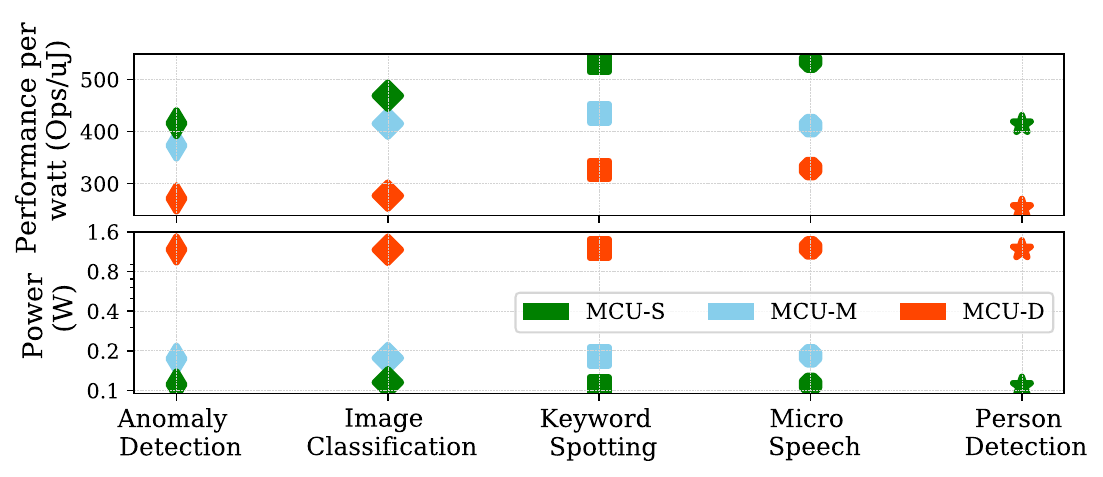}}
    \caption{Power and Performance per watt of MCUs.}
    \label{fig:PowerPerformancePerWattAll}
\end{figure}

Fig.~\ref{fig:power-performance-numbers} shows the latency/energy comparison for applications on different MCUs. We cannot measure the  power of MCU-B due to SmartPower2's restrictions, i.e., higher output voltage than what MCU-B can accept.  So, we only report energy numbers of other three MCUs.
MCU-D achieves the least model latency but the highest energy. Overall, the energy  is proportional to the inference latency for all the MCUs. {\em Anomaly detection} has the lowest latency/energy and {\em Image classification} has the highest latency/energy. This is due to the near-constant power  during inference. For instance, in Fig.~\ref{fig:PowerPerformancePerWattAll}, MCU-S power changes very little for different applications due to its relatively simple architecture. 

Fig.~\ref{fig:PowerPerformancePerWattAll} shows the performance per watt for three MCUs, where the data for MCU-B is not shown as we cannot measure its power. The performance per watt of MCUs always follows the order: MCU-S$>$MCU-M$>$MCU-D. The high performance per watt of MCU-S is due to its low frequency and small memory. Although the SRAM size of MCU-M is only half of that of MCU-S, its higher frequency causes lower performance per watt. Moreover, as Cortex-M4F is designed for high energy efficiency, its performance per watt is higher than that of Cortex-M7F (MCU-D). 

\textbf{Recommendation.} Executing the same model on an MCU with low computation/memory capability can achieve higher performance per watt.

\subsection{Frequency and memory bandwidth}
\label{subsection:frequency_bandwidth}
Next, we investigate the influence of hardware characteristics on inference performance. Table~\ref{tab:throughput_comparison_MCU} shows application throughput and hardware characteristics on MCUs, with MCU-S as the baseline. We observe that throughput is proportional to the bandwidth and frequency of MCU platforms. Note that MCU-D can dual-issue certain pairs of ALU instructions; so the throughput increase on MCU-D is twice its frequency increment compared to MCU-S. Also, as fully connected layers have more memory access per operation, higher memory bandwidth of MCU-D makes it achieve much lower latency. In contrast, convolution operations do not benefit much from high bandwidth. Therefore, among all the applications, {\em Image Classification}, which mainly consists of convolution layers, attains the best relative throughput on MCU-B and the worst relative throughput on MCU-D. Since the performance of {\em Image Classification} is more limited by compute capability instead of memory bandwidth, the impact of MCU memory bandwidth on application performance is weakened.  
\begin{table}[htbp]
    \centering
    \caption{Application throughput and hardware characteristics comparison on MCUs (Taking numbers on MCU-S as baseline)}
        \begin{tabular}{ccccc}
        \hline
            &\textbf{MCU-M}&\textbf{MCU-B}&\textbf{MCU-D}\\
            \hline
            Anomaly Detection&1.43&1.64&6.89\\
            \hline
            Micro Speech&1.27&1.53&6.67\\
            \hline
            Keyword Spotting&1.38&1.84&6.82\\
            \hline
            Person Detection&\XSolidBrush&1.77&6.63\\
            \hline
            Image Classification&1.36&\textbf{2.14}&\textbf{5.99}\\
            \hline
            \textbf{Average Throughput}&1.36&1.78&6.60\\
            \hline
            \textbf{Processor Frequency}&1.25&1.50&3.38\\
            \hline
            \textbf{Memory Bandwidth}&1.30&1.32&3.75\\
        \hline
        \end{tabular}
        \label{tab:throughput_comparison_MCU}        
\end{table}

\textbf{Recommendation.} Neural networks with more fully connected or DSC layers are better suited for MCUs with large memory bandwidth. Models with more convolutions are better suited for MCU platforms with high frequency.

\subsection{Memory Limitation}
\label{subsection:hardware_memory_limitation}

The ability of an MCU platform to support a specific ML application is dependent on memory requirements. Specifically, the application's working memory and executable binary have to reside in the platform's SRAM and Flash memory, respectively. We investigate these limitations with TFLM framework, the representative runtime for DNN on MCU.

The TFLM model's working memory mainly comprises intermediate tensors and persistent buffers, while generated binary includes model weights, graph definition, and operator implementation code~\cite{banbury2021micronets}. To further explore factors affecting model working memory and binary size, we select three models, MicroNet-KWS-L, MicroNet-KWS-M, and MicroNet-KWS-S for {\em Keyword Spotting}~\cite{banbury2021micronets}. 
Their specifications, including layer composition, TFLite model size, and \#MACC, are listed in Table \ref{tab:micronet-kws}. We choose similar model structures and operators to eliminate the influence of software libraries on memory requirements for different operation implementations. We report the inference performance and memory requirements on MCU-S and MCU-D, considering memory availability and processor diversity as  shown in Table \ref{tab:micronet-kws_MCUs}.  

\begin{table}[htbp]
    \centering
    \caption{Three MicroNet models for Keyword Spotting}
    \resizebox{\linewidth}{!}{
        \begin{tabular}{crrrr}
        \hline
            &\textbf{MicroNet-KWS-L}&\textbf{MicroNet-KWS-M}&\textbf{MicroNet-KWS-S}\\
            \hline
            Model&659 KB&182 KB&115 KB\\
            \hline
            Arena&205 KB&105 KB&70 KB\\
            \hline
            Layers&Conv+7 DSC+FC&Conv+5 DSC+FC&Conv+5 DSC+FC\\
            \hline
            \#MACC&65714k&15556k&8327k\\              
        \hline 
        \end{tabular}
    }
    \label{tab:micronet-kws}        
\end{table}

\begin{table}[htbp]
    \centering
    \caption{Three MicroNet model inference on two MCUs}
        \begin{tabular}{ccrr}
        \hline
            \multicolumn{2}{c}{\textbf{Applications}}&\textbf{MCU-S}&\textbf{MCU-D}\\
            \hline
            \multirow{3}{0.13\textwidth}{\centering \textbf{MicroNet-KWS-L}}&Working Memory&254 KB&215 KB\\
                \cline{2-4}
                &Binary&793 KB&738 KB\\
                \cline{2-4}
                &Latency&3790 ms&566 ms\\
            \hline
            \multirow{3}{0.13\textwidth}{\centering \textbf{MicroNet-KWS-M}}&Working Memory&152 KB&112 KB\\
                \cline{2-4}
                &Binary&316 KB&261 KB\\
                \cline{2-4}
                &Latency&1118 ms&172 ms\\
            \hline
            \multirow{3}{0.13\textwidth}{\centering \textbf{MicroNet-KWS-S}}&Working Memory&116 KB&76 KB\\
                \cline{2-4}
                &Binary&248 KB&193 KB\\
                \cline{2-4}
                &Latency&646 ms&99 ms\\
        \hline
        \end{tabular}
        \label{tab:micronet-kws_MCUs}        
\end{table}
    
\textbf{Working Memory.} 
TFLM defines a contiguous memory array called ``arena" for intermediate model results and persistent variables. 
During initialization, TFLM sets up the memory by overlapping memory allocations that are not simultaneously needed during the same operator evaluation. Therefore, the arena size is the maximal working memory needed by intermediate tensors during operator execution. It is determined by the model width, i.e., the layer with most inputs, weights, and activations. 
Hence, 
if the SRAM size is greater than the tensor arena size, the application performance cannot be improved further.  
Table \ref{tab:micronet-kws} lists the arena size of three MicroNets models, which are close to the SRAM size needed on MCU-D in Table \ref{tab:micronet-kws_MCUs}. A balanced model structure design in NAS may overcome the SRAM size limitation while maintaining sufficient accuracy. Some recent works apply operator reordering~\cite{liberis2019neural} or memory swapping~\cite{miao2021enabling} to reduce working memory or virtually expand working memory, respectively. 

\textbf{Executable Binary.} To further minimize the binary size to support model inference on MCUs with constrained flash memory, we investigate the composition of the executable binary and its relationship with the model structure. 
The generated binary is composed of the TFLite model and operator implementation code. 
Model structures like model depth and weight count are relative to the TFLite model size. The operator implementation code size is determined by model operator type and count. As shown in Table \ref{tab:micronet-kws_MCUs}, after excluding the TFLite model size, the remaining binary size is about 79 KB for all three models on MCU-D. This is the implementation code size for convolutions, depthwise separable convolutions, and fully connected operations. Compared with MCU-D, there is an extra 40 KB and 55 KB overhead in working memory and executable binary, respectively, on MCU-S. This is due to Arduino Nano 33 BLE Sense platform-specific overhead.  

\textbf{Inference Latency.} As the TFLite model is a composition of model graph definition and model weights, its size can, to some extent, reflect the inference latency. For the three models based on the same structure, model inference latency is proportional to the TFLite model size. For instance, the TFLite model size of MicroNet-KWS-L is 5.75 times that of MicroNet-KWS-S. Accordingly, the inference runtime of the large model on MCU-S is 5.86 times that of the small model, while on MCU-D, this number is 5.71. Compared with \#MACC, TFLite model size is a better indicator of inference latency. However, for five applications with different structures in Table~\ref{tab:model_description}, we do not observe an obvious relationship between latency and TFLite model size. 

{\bf Recommendation.} Balanced neural network model structure choice during NAS such that the largest layer can fit into on-chip memory is important for memory-constrained devices.

\section{Performance Prediction}
\label{section:performancePrdiction}
We now present an analytical approach for MCUs to predict the latency of each layer by taking into account the impact of the model, software, and hardware. 
Prior works such as~\cite{banbury2021micronets} assume that the model operation count can be regarded as an indicator for model performance. However, our evaluations found that a full-stack analysis is necessary for accurate prediction by capturing the impact of model parameters, software optimizations, and hardware characteristics.

\subsection{Analytical Approach}
\begin{figure}[htbp]
    \centerline{\includegraphics[width=1\linewidth]{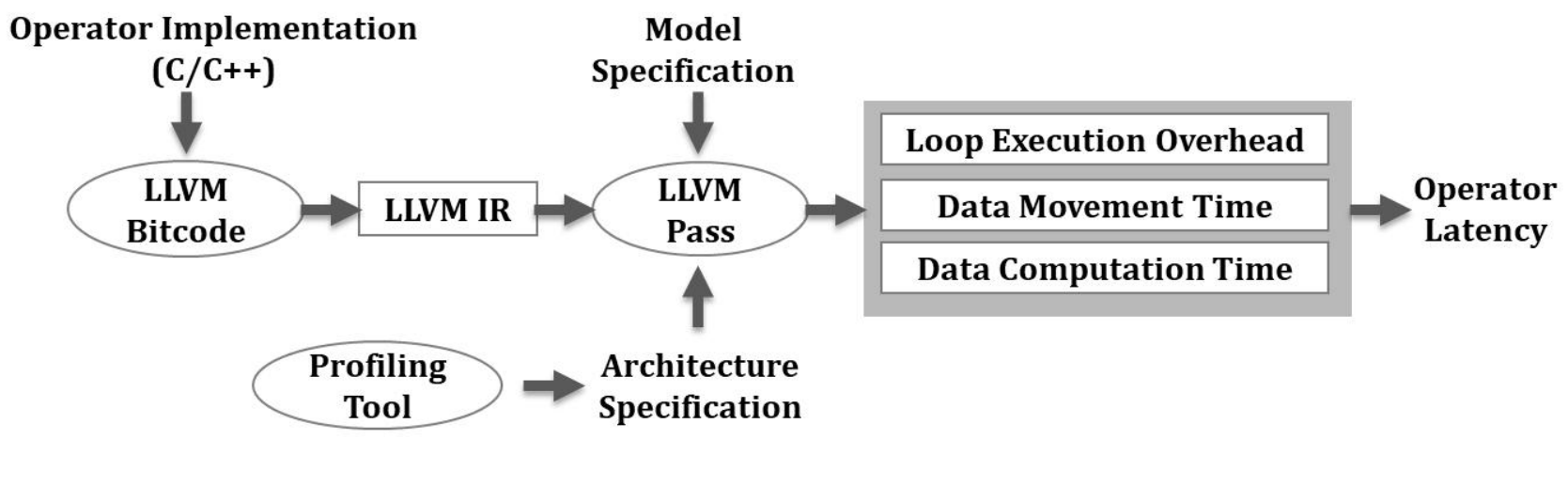}}
    \caption{Analytical approach for performance prediction.}
    \label{fig:PerformancePredictionModelMCU}
\end{figure}

Our analytical approach estimates inference time by estimating different types of assembly instructions executed during model inference. As shown in Fig.~\ref{fig:PerformancePredictionModelMCU}, we take in {\em Model Specification}, {\em Operator Implementation}, and {\em Architecture Specification} as inputs. We first generate the Low-Level Virtual Machine (LLVM) bitcode of operator implementation and obtain the corresponding intermediate representation (IR). Then based on IR, our LLVM pass counts assembly instructions for {\em Data Movement} and {\em Data Computation}, and loop iterations. Next, based on these data and architectural specifications, we estimate the inference time for {\em Data Movement}, {\em Data Computation}, and {\em Loop Execution} during layer inference. This estimated time constitutes the main latency of layer inference. 

\textbf{Inputs.}  \circled{1} {\em Model Specification.} Layer parameters such as filter height and width describe the layer structures, which offer us a high-level understanding of the compute and memory needs of the layers without any software optimization. 
\circled{2} {\em Operator Implementation.} CMSIS-NN kernels provide different optimization implementations for different operators. As mentioned in section~\ref{subsection:direct_convolution}, convolutions need the IM2COL process to align data for matrix multiplication, while fully connected layers do not need this extra input expansion. 
\circled{3} {\em Architecture Specification.} With pipeline processing, Cortex-M4 CPUs can execute one ALU instruction or a normal SIMD instruction each cycle. The integer-only model implementation based on CMSIS-NN kernels may need memory read or write instructions with different memory access sizes, like \textit{LDR}, \textit{STR}, \textit{LDRSB}, \textit{STRB}, etc. Moreover, based on different destinations, memory read/write instructions require different execution cycles. For instance, on MCU-S, reading four bytes from flash memory needs five cycles, while writing one byte to SRAM needs one cycle. Therefore, both memory access size and data location of the memory access instructions should be considered. Profiling tools, like J-Trace Pro Debug Probe, can determine the number of execution cycles of these assembly instructions on a given MCU device.

\textbf{Operator Latency.} We divide operator latency into the following three parts. 
\circled{1} {\em Loop Execution Overhead.} The number of control instructions required to manage the entire workflow correlates with the layer parameters. The layer with more computation requiring more GEMM conversion will cause more overhead from loop execution. We assume five cycles of overhead per iteration based on experimental evaluation, including the overhead of necessary jump instructions and potential instruction reloading under failed branch speculation. 
\circled{2} {\em Computation Time.} There is a high correlation between IR and actual assembly instructions for computation operations. We directly use the IR instructions count for computation time. 
\circled{3} {\em Data Movement Time.} We consider two data movement operations, loading or storing model weights, activations and loading or storing other variables defined in the program. The memory read/write instructions to load or store model weights, activations are easily recognized through data size and location, which are clearly shown in IR. In contrast, for the load/store of remaining variables, the LLVM IR cannot represent the process of loading these variables from SRAM to registers; so, we count the IR instruction \textit{PHI} to estimate this overhead of data transmission. One \textit{PHI} instruction corresponds to one variable load instruction from SRAM.


\subsection{Performance Prediction Results}

\begin{figure}[h]
    \centering
    \includegraphics[width=\linewidth]{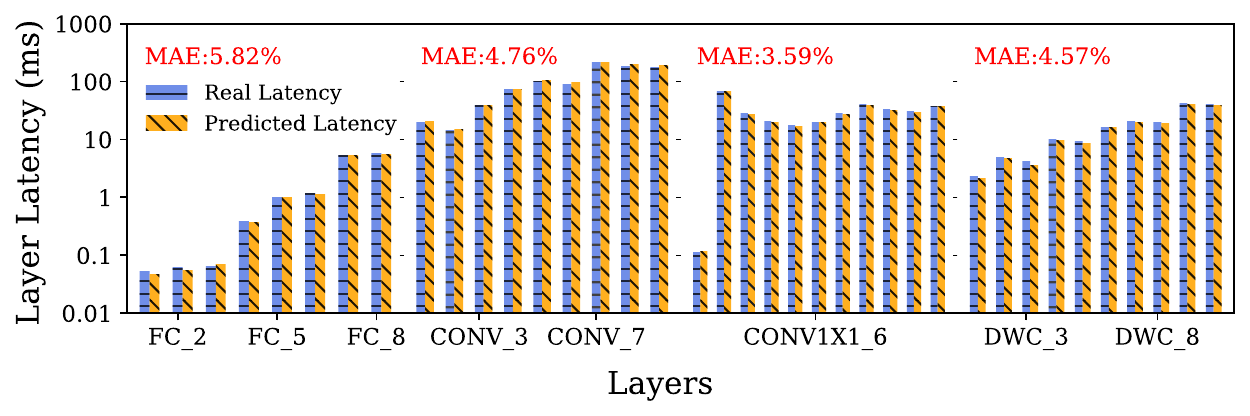}
    \caption{Predicted and real latency of layers on MCU-S.}
    \label{fig:LayerPrediction_Arduino}
\end{figure}

Fig.~\ref{fig:LayerPrediction_Arduino} shows the predicted and actual latency for FC, Conv, Conv1x1, and DWC layers on MCU-S. Clearly, our analytical approach can accurately predict layer inference latency. The absolute difference between predicted and actual latency is insignificant as shown in the Figure; the relative difference is 12.24\% across all layers, and the mean absolute error (MAE) is 4.59\%. We do not present results for {\em softmax}, {\em add}, and {\em reshape} operators as they occupy only a tiny fraction of model inference latency.

Besides, the predicted latency shows a similar trend to actual latency for all layers. Moreover, the predicted error decreases for large FC, Conv, and DWC layers with more computation. 
For instance, for the FC layer with a predicted error of 5.31\%, its actual latency is 63 $\mu$s while the predicted one is 66 $\mu$s. This slight difference in absolute latency causes a high relative prediction error. For larger layers, the relative prediction error is very low. 
All these observations reveal the effectiveness and robustness of our proposed performance prediction model. 

\begin{figure}[htbp]
    \centerline{\includegraphics[width=\linewidth]{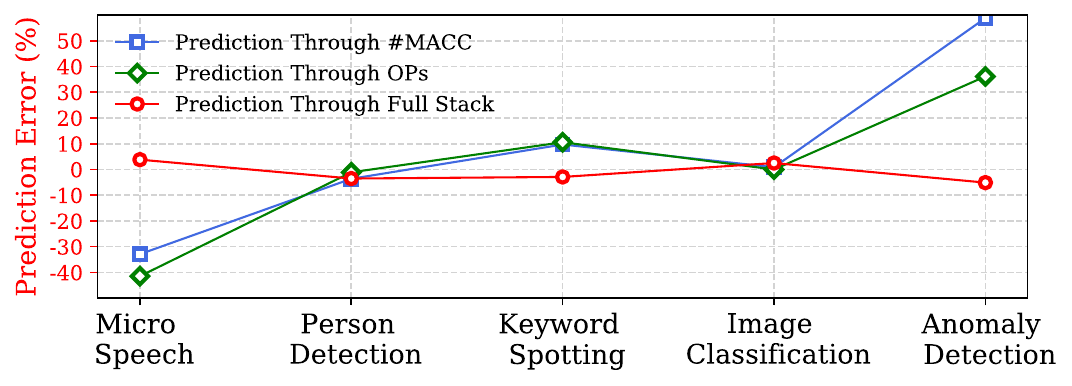}}
    \caption{Model prediction comparison on MCU-S.}
    \label{fig:ModelPrediction_Arduino}
\end{figure}

Based on predicted layer latency, we calculate model latency for five applications, with prediction error shown in Fig.~\ref{fig:ModelPrediction_Arduino}. Among five applications, the maximal difference between predicted and actual latency is 5.12\%, and MAE is 3.57\%. Fig.~\ref{fig:ModelPrediction_Arduino} also shows prediction error percentage based on model \#MACC or operation count (OPs). We obtain this number through linear regression analysis between model latency and \#MACC or OPs. It is observed that our prediction considering influence from full stack provides far more accurate results.  

\section{Conclusion}
In this paper, we present a systematic full-stack power-performance analysis of TinyML applications on multiple MCU platforms. Our analysis is performed from three different perspectives: the abstraction layers, the optimization library, and the hardware architectural impact. These observations provide insights and suggestions for NAS and system optimizations, as shown in each section. Finally, based on the analysis, we design a full-stack model to quickly estimate the latency of ML models accurately.

\section*{Acknowledgment}
This work was partially supported by the National Research Foundation, Singapore, under its Competitive Research Programme Award NRF-CRP23-2019-0003 and Singapore Ministry of Education Academic Research Fund T1 251RES1905.


\end{document}